\documentclass[conference, a4paper]{IEEEtran}

\usepackage{cite}
\usepackage{amsmath,amssymb,amsfonts}
\usepackage{algorithmic}
\usepackage{graphicx}
\usepackage{textcomp}
\usepackage{xcolor}
\usepackage[bottom]{footmisc}

\usepackage{balance}

\usepackage[linesnumbered,ruled,vlined]{algorithm2e}

\usepackage{amsmath}
\usepackage{amsthm}
\usepackage{amssymb, color}
\usepackage{epsfig} 

\newtheorem{lemma}{Lemma}
\newtheorem{sublemma}{Lemma}[lemma]

   \makeatletter
    \newcommand{\linebreakand}{%
      \end{@IEEEauthorhalign}
      \hfill\mbox{}\par
      \mbox{}\hfill\begin{@IEEEauthorhalign}
    }
    \makeatother

\begin{document}

\title{Convergence Rate Maximization for Split Learning-based Control of EMG Prosthetic Devices}

\author{\IEEEauthorblockN{1\textsuperscript{st} Matea Marinova}
\IEEEauthorblockA{\textit{Faculty of Electrical Engineering and} \\
\textit{Information Technologies} \\
\textit{Ss. Cyril and Methodius University} \\
Skopje, R. North Macedonia \\
mateam@feit.ukim.edu.mk}
\and
\IEEEauthorblockN{2\textsuperscript{nd} Daniel Denkovski}
\IEEEauthorblockA{\textit{Faculty of Electrical Engineering and} \\
\textit{Information Technologies} \\
\textit{Ss. Cyril and Methodius University} \\
Skopje, R. North Macedonia \\
danield@feit.ukim.edu.mk}
\and
\IEEEauthorblockN{3\textsuperscript{rd} Hristijan Gjoreski}
\IEEEauthorblockA{\textit{Faculty of Electrical Engineering and} \\
\textit{Information Technologies} \\
\textit{Ss. Cyril and Methodius University} \\
Skopje, R. North Macedonia \\
hristijang@feit.ukim.edu.mk}
\and
\linebreakand
\IEEEauthorblockN{4\textsuperscript{th} Zoran Hadzi-Velkov}
\IEEEauthorblockA{\textit{Faculty of Electrical Engineering and} \\
\textit{Information Technologies} \\
\textit{Ss. Cyril and Methodius University} \\
Skopje, R. North Macedonia \\
zoranhv@feit.ukim.edu.mk}
\and
\IEEEauthorblockN{5\textsuperscript{th} Valentin Rakovic}
\IEEEauthorblockA{\textit{Faculty of Electrical Engineering and} \\
\textit{Information Technologies} \\
\textit{Ss. Cyril and Methodius University} \\
Skopje, R. North Macedonia \\
valentin@feit.ukim.edu.mk}
}
\maketitle

\begin{abstract}
Split Learning (SL) is a promising Distributed Learning approach in electromyography (EMG) based prosthetic control, due to its applicability within resource-constrained environments. Other learning approaches, such as Deep Learning and Federated Learning (FL), provide suboptimal solutions, since prosthetic devices are extremely limited in terms of processing power and battery life. The viability of implementing SL in such scenarios is caused by its inherent model partitioning, with clients executing the smaller model segment. However, selecting an inadequate cut layer hinders the training process in SL systems. This paper presents an algorithm for optimal cut layer selection in terms of maximizing the convergence rate of the model. The performance evaluation demonstrates that the proposed algorithm substantially accelerates the convergence in an EMG pattern recognition task for improving prosthetic device control.

\end{abstract}

\begin{IEEEkeywords} 
Convergence rate maximization, electromyography, optimal cut layer, prosthetic control, split learning
\end{IEEEkeywords}

\section{Introduction}
Centralized Machine Learning (ML) involves transmitting a vast amount of raw data, which may cause both increased latency and potential network congestion. Distributed learning techniques, such as Federated Learning (FL) and Split Learning (SL), actively address these issues by conducting distributed training without transmission of raw data between the clients and the server [\ref{distr}]. Therefore, these distributed learning approaches are adequate for scenarios involving time-varying and resource-constrained systems.

FL is a distributed learning paradigm in which clients perform collaborative training of a global model. Each of the participating clients maintains a copy of the complete model and trains it using the local dataset. Subsequently, the central entity receives these local models in order to generate a global one. This implies that in FL, the server performs low-demanding tasks in terms of computation, which leads to suboptimal resource utilization, assuming that it has higher computational capabilities compared to the clients [\ref{fl}].

SL is another distributed learning strategy that has emerged as an alternative, which successfully resolves the considered challenges faced by FL. In SL, the model is partitioned into separate client-side and server-side segments at a layer known as cut layer or split layer. By having the server process a larger fraction of the model, its resources are employed more efficiently [\ref{sl1}]. Furthermore, the training involves transmission of the cut layer' activations and gradients only rather than the entire model parameters. This fundamental property provides further privacy preservation compared to FL by restricting the entities' access only to their designated sub-networks. As such, SL is a promising approach within domains where the involved clients require a higher degree of privacy, for example, in healthcare [\ref{health2}].

Employing SL is also suitable in scenarios featuring clients with limited computational resources.  The wearable devices utilized in both electromyography (EMG) based prosthetic control and Human Activity Recognition (HAR) belong to this category. Specifically, a significant issue caused by the low processing power of EMG controlled prostheses is their inability to incorporate large models. This property makes the application of deep learning and FL impractical in such systems. Moreover, implementing small models does not satisfy the strict requirements regarding the recognition accuracy.  Deployment of SL can potentially solve the specified problem by assigning only a small model segment to the prosthetic device.

One of the primary challenges in SL is choosing the cut layer, as this decision can have a profound impact on the effectiveness of the deployed ML model. In real-world deployments, system resources are limited and exhibit frequent variations. Hence, SL must employ dynamic cut layer selection, that optimizes the learning capabilities of the underlying model, under specific constraints. This paper does not only demonstrate the viability of incorporating SL in EMG controlled prosthetic technology, but also develops a cut layer selection algorithm for maximizing the convergence rate, considering the limited and time-varying system resources.

The paper is structured as follows. Section II provides related work regarding deep learning based HAR solutions, mostly focusing on EMG pattern recognition for improving prosthetic control, and algorithms for cut layer selection in SL systems. In Section III, a detailed overview of the considered SL system model is presented. Section IV gives a thorough explanation of the proposed optimal cut layer selection algorithm. Section V features a performance comparison between the proposed and the benchmark algorithm in an EMG pattern recognition task. Section VI concludes the paper.

\section{Related Work}

According to [\ref{EMG-PR}], EMG pattern recognition based control of prosthetic devices relies on the hypothesis that EMG patterns contain significant information about the intended movements.
Deep learning techniques can effectively address some weaknesses of conventional machine learning approaches in EMG pattern recognition [\ref{EMG-PR}] and HAR, such as their dependence on statistical features, and domain knowledge of the system. Furthermore, traditional techniques incorporate shallow features unable to recognize complex activities. Deep learning mitigates these setbacks through feature learning directly from raw sensors' data without feature engineering, and domain knowledge as well [\ref{perscomm}]. Hence, deep neural networks are becoming increasingly favored in these domains, especially Convolutional Neural Networks (CNNs), which are known for their ability to capture complex features by combining previously learned simpler ones. 

These studies [\ref{zebin}], [\ref{emg}], [\ref{NN}] demonstrate that CNN models exhibit superior performance in both EMG pattern recognition and HAR-specific tasks compared to statistical machine learning techniques, such as K-Nearest Neighbors (KNN), Support Vector Machine (SVM), Random Forest (RF), etc. In addition, [\ref{zebin}] provides a proof of concept for implementation of deep learning models on edge devices (e.g. smartphones). In [\ref{DL-HAR}], a comparison of deep neural network architectures is provided, which shows that CNNs outperform Long Short-Term Memory (LSTM), Bidirectional LSTM (biLSTM) and Deep Belief Networks (DBNs), and their results are competitive with the scores of Gated Recurrent Unit (GRU) networks in the HAR domain. Furthermore, the results reveal that CNN architectures are adequate for tasks in resource-constrained environments, regarding both speed and memory requirements. The prospect of employing deep learning solutions in the field of EMG-based prosthetic device control implies that SL can be incorporated within the specified domain as well. Moreover, this research direction  has yet to be explored, since only [\ref{birnn}] combines SL and HAR, using large Bidirectional Recurrent Neural Network (Bi-RNN) for indoor activity recognition.

The effectiveness of SL solutions is intricately connected to the cut layer decision. Attempts in the existing literature have been made in terms of creating algorithms for cut layer selection. The prevalent strategy relies on implementing the brute-force method [\ref{ares}],[\ref{jsac}]. The authors in [\ref{hivemind}] propose an efficient algorithm for multi-split scenarios, however its relevance diminishes when considering single-split SL systems, since the graph transformations and pruning are not applicable. In [\ref{barg}], the Kalai–Smorodinsky bargaining solution for finding the optimal cut layer is employed, which is an advanced game theory algorithm, based on assumption approximations. As such, it is not only complex, but may also result in suboptimal solutions. In [\ref{epsl}], to solve the cut layer selection problem, the authors implement the branch-and-bound algorithm. However, bounding techniques possess a critical role in its performance. Additionally, this algorithm demonstrates high computational complexity when used for solving large problems.

\section{System Model}

\begin{figure}[b]
\vspace{-3mm}
\centering
\includegraphics[scale=0.45]{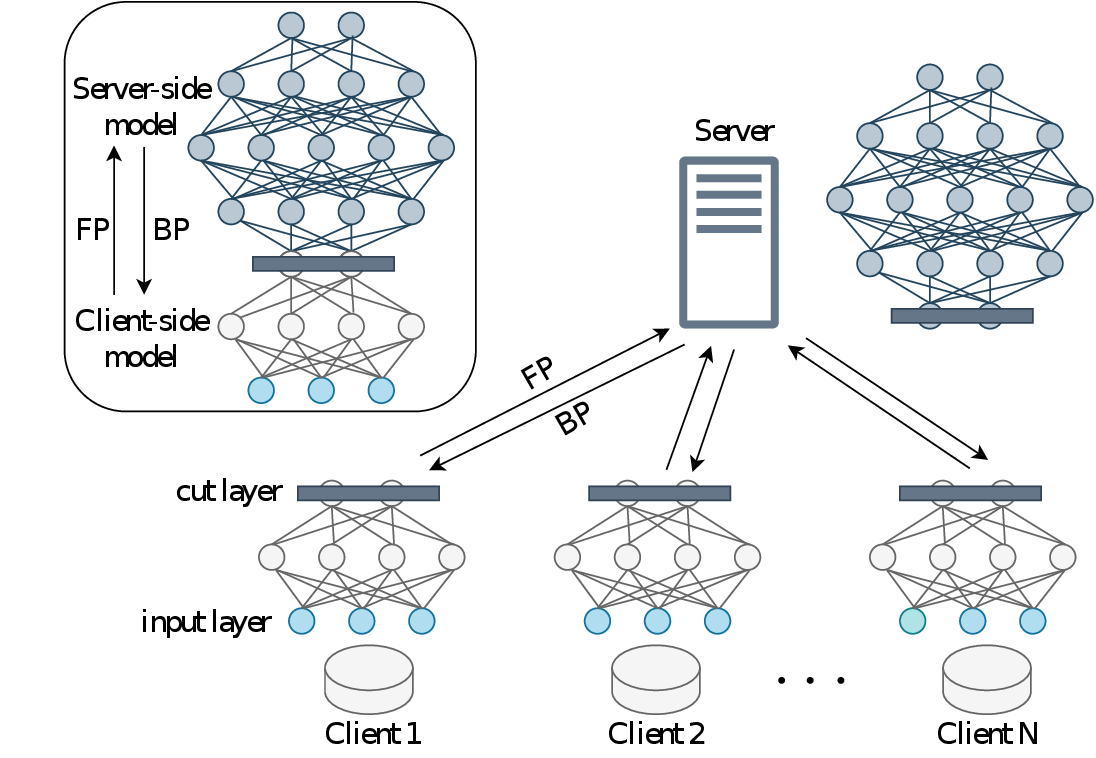} \vspace{-2mm}
\caption{Generic split learning architecture composed of N clients, which contain the first three layers of the neural network; the remaining four layers are assigned to the server}
\vspace{-2mm}
\label{architecture}
\end{figure}

The system model assumes a SL setup composed of one server and multiple heterogeneous clients with limited computation and communication resources, such as the SL configuration illustrated in Fig. \ref{architecture}. EMG controlled prostheses fall into this category of resource-constrained devices. The learning is performed in collaborative settings and sequential manner.

In SL, prior to the training procedure, the model parameters are initialized randomly, and the neural network is divided into two segments, with one segment assigned to the client, while the other segment resides on the server side. For each batch of samples, the procedure involves Forward Propagation (FP) and Backpropagation (BP). In that regard, during FP the samples are processed by the client up to the cut layer, which subsequently sends the resulting cut layer's outputs (i.e. smashed data) to the server. Through this approach, the server can proceed further with the FP process using solely the smashed data, without requiring access to the client's raw data. Then, in BP the gradients are computed with respect to the considered loss function. The smashed data's gradients are transmitted to the client, which continues and terminates the BP process accordingly. 

Additionally, weight synchronization is conducted between the clients participating in the training process. This implies that before the start of the client's epoch, it receives the client-side model parameters from the server, which were updated during the training epoch of the previous client. It is possible to implement SL without weight synchronization, however this variant does not guarantee that the model will converge [\ref{comparison}]. 

The entire SL procedure is illustrated in algorithm \ref{alg}. It is assumed that in one SL round each client trains on its local dataset for a single epoch, before the server proceeds to train the next client's model. In algorithm \ref{alg}, an epoch is described with lines 8-14, where $D_k$ and $B_k$ denote the client's dataset size and batch size, respectively.

\RestyleAlgo{ruled}
\begin{algorithm}[b!]
\caption{Split learning procedure}\label{alg}
\begin{small}
\For {$training$ $round$ $t = 1,2,...,T$}{
   \For {$client$ $c = 1,2,...,N$} {
        Divide the neural network into client-side and server-side segments;\\
        \eIf{t=1 \textbf{and} c=1}{
        Randomly initialize the parameters;}{
        The server sends an appropriate model segment to client $c$ for weight synchronization;\\
        } 
        \For {$each$ $batch$ $b = 1,2,...,\frac{D_k}{B_k}$} {
            The client performs FP on the client-side model;\\
            The client sends the smashed data to the server;\\
            The server performs FP on the server-side model;\\
            The server performs BP on the server-side model;\\
            The server transmits the smashed data's gradients to the client;\\
            The client performs BP on the client-side model;\\
        }
        \If{$t \neq  T$ \textbf{or} $c \neq  N$}{
            Client $c$ sends its model parameters to the server, enabling weight synchronization for the subsequent client;\\
        } 
   }
}
\end{small}
\end{algorithm}

The system model is built on the premise of wireless transmission over reciprocal uplink and downlink fading channels. As such, the specified model considers the inevitable channel state variations inherent to wireless channels. Furthermore, the fluctuations in the computational resources of the participating entities are considered as well, since they may occur due to other applications running in parallel. However, during a single epoch, system stability is assumed, i.e. the specified system resources do not exhibit any fluctuations.

This study focuses on selecting the optimal cut layer in order to accelerate the convergence rate of the model, i.e. to minimize the overall training delay composed of four main components: i) the computation time of the client ($\tau_k$), ii) the computation time of the server ($\tau_s$), iii) the time required for transmitting activations/gradients during FP/BP ($t_0$), and iv) the weight synchronization transmission delay ($t_p$). To simplify the analysis while still preserving its generality, it is assumed that the forward propagation and backpropagation delays ($\tau_k$, $\tau_s$, $t_0$) are equal [\ref{jsac}],[\ref{comparison}],[\ref{IoTSL}]. Thus, for client's dataset size $D_k$ and batch size $B_k$, the training delay per epoch is given by:
\begin{equation} \label{totalDelay}
    T=\frac{2D_k}{B_k}(\tau_k+t_0+\tau_s)+2t_p, \vspace{-1mm}
\end{equation}
where each of the components depends on the position of the split layer. It should be noted that at the beginning of the learning process, the first client does not receive any parameters from the server. However, this exception applies only to the first training round. Another exception occurs in the final round, i.e. the last client does not send its model parameters to the server for weight synchronization.

The device-side and server-side computational loads for one sample can be defined as: \vspace{-2mm}
\begin{subequations} \label{comp_loads}
\begin{align}
& L_k(i)=\sum_{j=1}^{i}l(j)\label{Lk}\\
& L_s(i)=L_{total}-L_k(i), \label{Ls}
\end{align}
\end{subequations}
where $i$ is the position of the selected cut layer, and $l(j)$ denotes the computational load per sample for layer $j$, obtained by multiplying the number of outputs by the number of FLOPs needed for computing each output. Therefore, the computation delays of the client and the server are given by:
\begin{subequations} \label{comp_delays}
\begin{align}
& \tau_k(i)=\frac{L_k(i)B_k}{f_k} \label{tau_k}\\ 
& \tau_s(i)=\frac{L_s(i)B_k}{f_s} \label{tau_s}
\end{align}
\end{subequations}
where $f_k$ and $f_s$ are the computing speeds of the client and the server, respectively, measured in FLOPS (floating-point operations per second).

Transmitting $N_k(i)$ activations/gradients of the split layer $i$  for $B_k$ samples per batch through a wireless channel with transmission rate $R$, introduces the following transmission delay: 
\begin{equation} \label{t0}
    t_0(i)=\frac{N_k(i)B_k}{R}.
\end{equation}
The delay induced by the transmission of client-side model parameters between the server and the client for weight synchronization is: \vspace{-1mm}
\begin{equation} \label{tp}
    t_p(i)=\frac{N_c(i)}{R},
\end{equation}
where $N_c(i)=\sum_{j=1}^{i}N_p(j)$ represents the cumulative function of the client-side model parameters, and $N_p(j)$ denotes the number of parameters in a given layer $j$.

It is evident that choosing a split layer from the initial layers in the neural network is computationally more efficient, under the assumption that the server has superior computing capability. Moreover, these split layer candidates result in lesser communication overhead caused by the weight synchronization. Conversely, the gradual activation size reduction in a convolutional neural network, indicates that opting for a split layer from the latter part of the network is more beneficial in terms of reducing the communication overhead stemming from the transmission of activations and gradients. This signifies that identifying the optimal cut layer in a specific time frame, is dependent on the interrelation between the available fluctuating system resources and the architectural aspects of the considered neural network. Given these factors, the following section outlines the proposed algorithm for optimal cut layer selection.

\section{Optimal Cut Layer Selection Algorithm}
This section elaborates on the proposed Optimal Cut Layer Selection Algorithm (OCLA), which consists of two separate phases. The first phase is conducted offline, while the second one operates in online mode.

\begin{figure}[b]
\centering
\vspace{-5mm}
\includegraphics[scale=0.54]{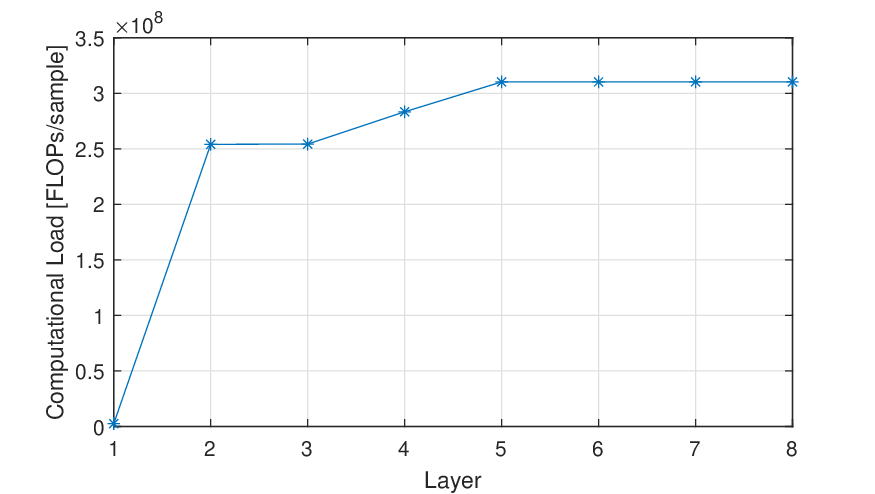} \vspace{-2mm}
\caption{Cumulative client-side computational load per layer} \vspace{-1mm}
\label{graph_compload}
\end{figure}

\begin{figure}[h]
\centering
\includegraphics[scale=0.53]{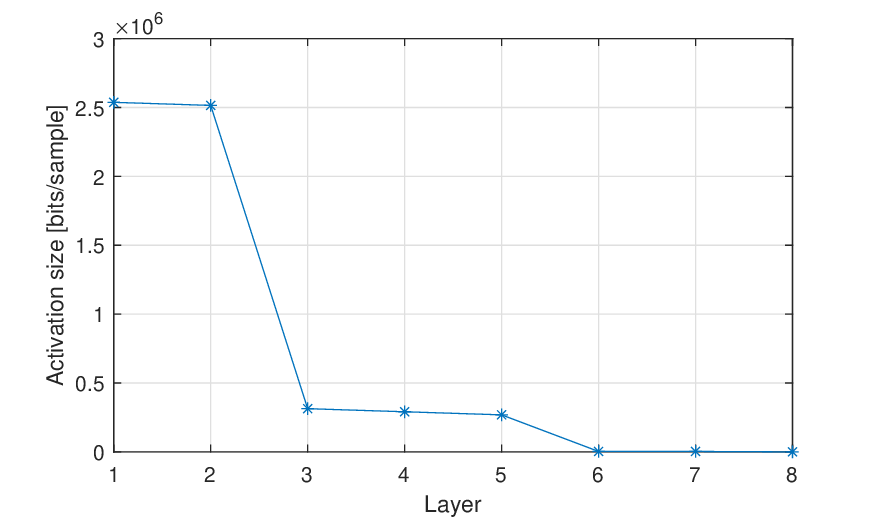} \vspace{-3mm}
\caption{Per-layer activation size}
\label{graph_act}
\vspace{-3mm}
\end{figure}

\begin{figure}[h]
\centering
\includegraphics[scale=0.53]{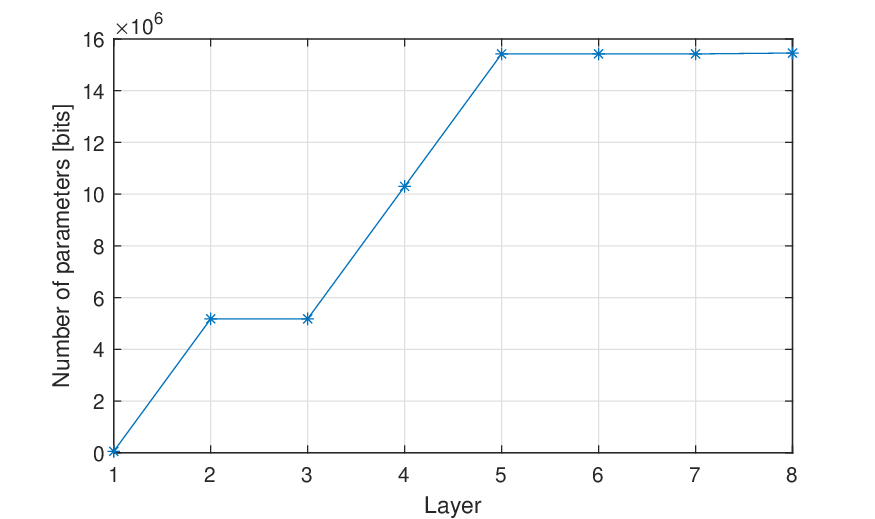} \vspace{-3mm}
\caption{Cumulative client-side model parameters per layer}
\label{graph_param}
\vspace{-3mm}
\end{figure}

A prerequisite for the offline phase of the algorithm is having a thorough understanding of the neural network architecture, as well as the size of the dataset that the model will be trained on.
In essence, for a given neural network, the pool of potential cut layers can be determined entirely based on the profiling functions\footnote{The illustrated profiling functions are derived from the neural network specified in Section V.} depicted in Fig. \ref{graph_compload}, Fig. \ref{graph_act} and Fig. \ref{graph_param}. This indicates that the elimination of inadequate candidates from the selection process is feasible without prior information about the SL system conditions, i.e. the available system resources. The specified pruning method is composed of two distinct steps.

It can be observed from Fig. \ref{graph_compload} that the client-side computational load is a cumulative function. As such, it dictates that only layers which satisfy progressively decreasing communication overhead function of the neural network are considered as potential candidates in the selection process. This analysis leads to the formulation of the first pruning step called profile function pruning, which eliminates the layer on position $i+1$ if it complies with the following condition: 
\begin{equation} \label{comm}
   \scalebox{1.02}{$N_k(i+1) + \frac{N_c(i+1)-N_c(i)}{D_k} \geq  N_k(i) , \ \forall i \in \{1,2,...,M-1\}.$}
\end{equation}
\begin{IEEEproof}
Please refer to Appendix A. \vspace{+2mm}
\end{IEEEproof}

In (\ref{comm}), M denotes the final layer in the considered neural network. This layer is always excluded from the selection process, since choosing it would imply that the entire model is assigned to the client, which contradicts the core definition of SL. This condition indicates that the activation size, depicted in Fig. \ref{graph_act}, affects to a greater extent the cut layer decision compared to the weight synchronization parameters. Moreover, it indicates that as the dataset size grows, the impact of these parameters (Fig. \ref{graph_param}) on the cut layer selection process further diminishes. 

The second pruning step is performed using the communication-computation trade-off function for consecutive layers, which is defined as: \vspace{-1mm}
\begin{subequations} \label{trade-off}
\begin{align}
& \Delta(j,j+1) = \frac{N_{k}(j)-N_{k}(j+1)-\frac{N_c(j+1)-N_c(j)}{D_k}}{L_{k}(j+1)-L_{k}(j)}\label{gain_plus} \vspace{+2mm}\\ 
& \Delta(j-1,j) = \frac{N_{k}(j-1)-N_{k}(j)-\frac{N_c(j)-N_c(j-1)}{D_k}}{L_{k}(j)-L_{k}(j-1)}, \label{gain_minus}
\end{align}
\end{subequations}
\ $\forall j \in \{1,2,...,P\}$, where P is the last layer in the pool of potential cut layers after completing the first pruning step.

For proper execution of OCLA, two modifications to the $\Delta$ function are applied. First, it is assumed that $\Delta(0,1) \rightarrow \infty$. This aligns with the observation that the first layer is the most efficient candidate regarding the processing aspect. It is important to consider it as a potential cut layer, since one of the main advantages of using the SL approach is exactly its capability to decrease the necessary computation performed by the resource-constrained clients. The second alteration involves adding a virtual final layer $P+1$ characterized by activation size $N_{k,v}(P+1)=0$, computational load $L_v(P+1)=0$ and number of parameters $N_{p,v}(P+1)=0$, which leads to $\Delta(P,P+1) < 0$.

The second step in the offline phase is performed by pruning the layers that do not satisfy the following condition:
\begin{equation} \label{pruning_gain}
    \Delta(j-1,j) > \Delta(j,j+1), \ \forall j \in \{1,2,...,P\}
\end{equation}
In fact, condition (\ref{pruning_gain}) ensures that the communication-computation trade-off function is a strictly decreasing one. It should be noted that this step might require to be executed more than once, until the desired behavior of $\Delta$ is achieved.

An important advantage of the discussed pruning steps is that they can be applied before using other algorithms for cut layer selection as well. Therefore, they are capable of decreasing substantially the computational complexity of the given algorithm, even when employed in conjunction with computationally inefficient methods such as exhaustive search.

After the pruning is completed, OCLA inspects whether a given layer fulfills the concluding condition for optimality, which is based on the available system resources. This condition stems from the fact that the optimal cut layer $n$ results in lower training delay compared to both of his adjacent layers $n-1$ and $n+1$:
\begin{subequations} \label{Tplusminus}
\begin{align}
& T(n)<T(n+1) \label{Tplus}\\
& T(n)<T(n-1),  \forall n \in \{1,2,...,K\} \label{Tminus}
\end{align}
\end{subequations}
where $T(n)$ represents the overall delay if $n$ is selected as the cut layer, and $K$ is the number of elements in the set of potential cut layers after the discussed pruning steps are performed.

\setcounter{lemma}{1}
\begin{sublemma}
Assume that the communication-computation trade-off between layer $n$ and layer $n+1$ is known and specified as in (\ref{gain_plus}), and that (\ref{Tplus}) always holds. If $n$ is the optimal cut layer, then the respective communication-computation trade-off is bounded by: 
\begin{equation} \label{left_delta}
    \Delta(n,n+1) < \frac{\beta R}{f_k}. \vspace{-1mm}
\end{equation}
\
\label{lemma1}  
\vspace{-3mm}\end{sublemma}
\begin{sublemma}
Assume that the communication-computation trade-off between layer $n$ and layer $n-1$, is known and specified as in (\ref{gain_minus}) and that (\ref{Tminus}) always holds. If $n$ is the optimal cut layer, then the respective communication-computation trade-off is bounded by: 
\begin{equation} \label{right_delta}
    \Delta(n-1,n) > \frac{\beta R}{f_k}. \vspace{-1mm}
\end{equation}
\label{lemma2}
\vspace{-3mm}\end{sublemma}
\begin{IEEEproof}
Please refer to Appendix B. \vspace{+2mm}
\end{IEEEproof}

The ultimate criterion, which if satisfied ensures the optimality of the considered cut layer candidate depending on the available system parameters $f_k$, $f_s$ and $R$, can be stated as:
\begin{equation} \label{optimality}
    \Delta(n,n+1) < \frac{\beta R}{f_k} <  \Delta(n-1,n), \ \forall n \in \{1,2,...,K\}
\end{equation}
 where $a=\frac{f_s}{f_k}$, and $\beta=\frac{a-1}{a}$ is the relative ratio between the computing speeds of the server and the client.

Based on the specified optimality condition, the final offline step includes creating a database which contains the split regions, defined by the communication-computation trade-off functions for the K remaining potential cut layers.

A key advantage of OCLA lies in its ability to determine all of the potential cut layers in the absence of information about the available system resources. Specifically, OCLA identifies the set of K remaining layers solely based on the neural network architecture and other input parameters, such as the
dataset size. However, only one layer satisfies the optimality condition (\ref{optimality}) for a given neural network and system parameters, and the subsequent online phase is dedicated to identifying it.

In essence, the profiling is an offline optimization step performed only once for a given neural network architecture and dataset size, and the obtained results are used during the online phase. The online part of OCLA consists of calculating the value of $\frac{\beta R}{f_k}$, and determining the split region from the database within which this value resides. By locating the split region, the optimal cut layer is successfully identified.

\section{Results}

The performance of OCLA is analyzed in a SL scenario where the training is performed sequentially between the server and N=10 heterogeneous clients. Specifically, OCLA is applied per each client's epoch, in order to accelerate the overall convergence rate. A detailed list of the simulation parameters is provided in Table \ref{parameters}. The analysis considers the time-varying inherent property of the computation and communication resources by modeling the transmission rate and the server-client computing speed ratio as random variables that follow folded normal distributions. The performance evaluation of OCLA features a comparison with a naive algorithm as the baseline approach, which persistently selects the same cut layer (e.g. the first layer), disregarding the impact of the neural network architecture in the cut layer selection process, and the variability of the system resources as well.

In order to achieve statistically accurate results, since the system parameters are considered to be random variables, a Monte Carlo simulation is performed as part of the evaluation process. Specifically, the simulation involves a particular number of iterations (I), within each iteration J samples are drawn from the folded normal distributions, and subsequently used for comparing both algorithms in terms of the given performance metrics. In this study, the comparative analysis is performed by using the performance gain, and the model's training accuracy and loss, as key performance metrics.

Our experiments are conducted using the dataset from [\ref{dataset}], which consists of EMG signals collected from 10 subjects with two EMG channels. The data collection process involved 6 trials, and within each trial the participants repeated movements classified in 10 different classes, i.e. flexion of each of the individual fingers, pinching between each finger and the thumb, and hand close. The collected data is partitioned into training and test datasets, by allocating five trials of each subject into the training dataset, and allocating the remaining trial to the test set. After performing the segmentation, each client's training dataset contains 9992 samples, and their respective test sets contain 1992 samples. During model training, the cross entropy loss and the Adamax optimizer are used. The specific 1D CNN model architecture employed for obtaining the results is described in [\ref{NN}], and shown in Table \ref{table}. It should be noted that that the input layer cannot be a cut layer in SL, as such its index is considered to be 0.
\begin{table}[t]
\caption{Simulation Parameters}
    \centering
    \begin{tabular}{p{0.1mm}p{4.8cm}|p{2.7cm}} \hline
         & {\textbf{Parameter}} & {\textbf{Value}} \\ \hline
         & Number of layers in the neural network & $M=8$ \\
         & Number of iterations  & $I=1000$ \\
         & Number of samples  & $J=300$ \\
         & Mean value of the computing speeds' ratio &  $\mathbb{E}[1-\beta]=0.03$ \\
         & Standard deviation of the computing speeds' ratio & $\sigma_{1-\beta}= 0.0003-0.015$ \\
         & Coefficient of variation of the computing speeds' ratio & $(1-\beta)_{cv}=0.01-0.5$ \\ 
         & Mean value of the transmission rate &  $\mathbb{E}[R]=20$ Mbps \\ 
         & Standard deviation of the tranmission rate  & $\sigma_{R}= 0.2-10$ Mbps \\
         & Coefficient of variation of the transmission rate & $R_{cv}=0.01-0.5$\\
         & Number of training rounds & $T=35$ \\
         & Number of epochs per round & $E=1$ \\
         & Number of clients & $N=10$ \\
         & Client's dataset size  & $D_k=9992$ \\ 
         & Batch size  & $B_k=100$ \\ \hline
    \end{tabular}
    \label{parameters}
\vspace{-2.5mm}
\end{table}
\begin{table}[b]
\vspace{-3mm}
\caption{Neural Network Architecture}
    \centering
    \begin{tabular}{ccccc}  \hline
         & {\textbf{Index}} & {\textbf{Layer Name}} & {\textbf{Output Size}} & {\textbf{Activation Function}} \\ \hline
         & 0 & Input layer & 800x2 & None \\
         & 1 & CONV 1 & 793x200 & ReLU \\
         & 2 & CONV 2 & 786x200 & ReLU \\
         & 3 & POOL 1 & 98x200 & None \\
         & 4 & CONV 3 & 91x200 & ReLU \\
         & 5 & CONV 4 & 84x200 & ReLU \\
         & 6 & Global Avg POOL & 1x200 & None \\
         & 7 & DROPOUT & 1x200 & None \\
         & 8 & FC & 10 & Softmax \\ \hline
    \end{tabular}
    \label{table}
\vspace{-2mm}
\end{table}

\begin{figure}[t]
\centering
\includegraphics[scale=0.46]{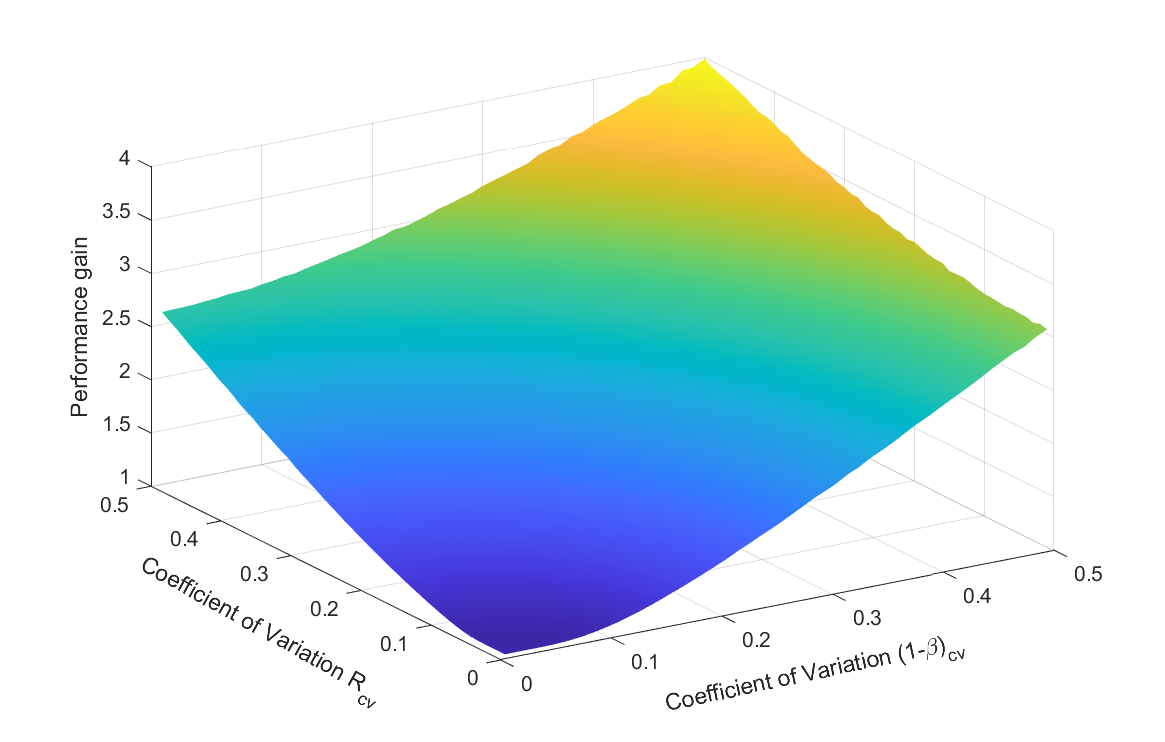} \vspace{-7mm}
\caption{Performance gain of OCLA regarding the coefficients of variations of the transmission rate $R_{cv}$ and the computing speeds' ratio $(1-\beta)_{cv}$; the baseline algorithm consistently selects layer 3} \vspace{-4mm}
\label{perf_gain}
\end{figure}

Fig. \ref{perf_gain} illustrates the performance gain of OCLA in comparison to the naive algorithm considering a scenario in which it consistently select layer 3 as a split layer. The performance gain of a given algorithm depends on the coefficients of variations $(1-\beta)_{cv}$ and $R_{cv}$ defined as:
\begin{subequations} \label{CV}
\begin{align}
(1-\beta)_{cv} &=\frac{\sigma_{1-\beta}}{\mathbb{E}[1-\beta]}\\ \vspace{+2mm}
R_{cv} &=\frac{\sigma_{R}}{\mathbb{E}[R]}.  \label{Ccv}
\end{align}
\end{subequations}
Therefore, the performance gain metric is determined by:
\begin{equation} \label{gain}
    gain(R_{cv},(1-\beta)_{cv}) = \frac{A_{OCLA}(R_{cv},(1-\beta)_{cv})}{A_{naive}(R_{cv},(1-\beta)_{cv})}.
\end{equation}
In (\ref{gain}), $A_{OCLA}$ and $A_{naive}$ denote the optimal cut layer selection rates of OCLA and the naive algorithm, respectively. The optimal cut layer selection rate is specified as: \vspace{-1mm}
\begin{equation} \label{acc}
    A = \frac{\sum_{i=1}^{P}I_A(y_i=\hat{y_i})}{P}, 
\end{equation}
where $y_i$ and $\hat{y_i}$ represent the optimal cut layer and the cut layer selected by the considered algorithm; $I_A(\cdot)$ is an indicator function; $P$ denotes the total number of predictions.

It is evident that for lower values of the coefficients of variations defined by (\ref{CV}), the performance gain of OCLA is lower. However, the gain substantially increases if at least one of the coefficients increase, or both. Specifically, for lower $(1-\beta)_{cv}$ and $R_{cv}$, the considered range of values for the system parameters is too small, and focused on the split region of layer 3. As such, they most frequently lead to optimality of cut layer 3, causing a higher accuracy of the baseline algorithm. Conversely, by widening the range of values of the observed system parameters, more layers emerge as optimal cut layers, i.e. more split regions are included, which contributes to a decrease in accuracy of the baseline algorithm. This property holds significant importance, since non-deterministic time-varying parameters exhibit higher coefficients of variation within practical real-world environments. 

\begin{figure}[h]
\centering
\includegraphics[scale=0.49]{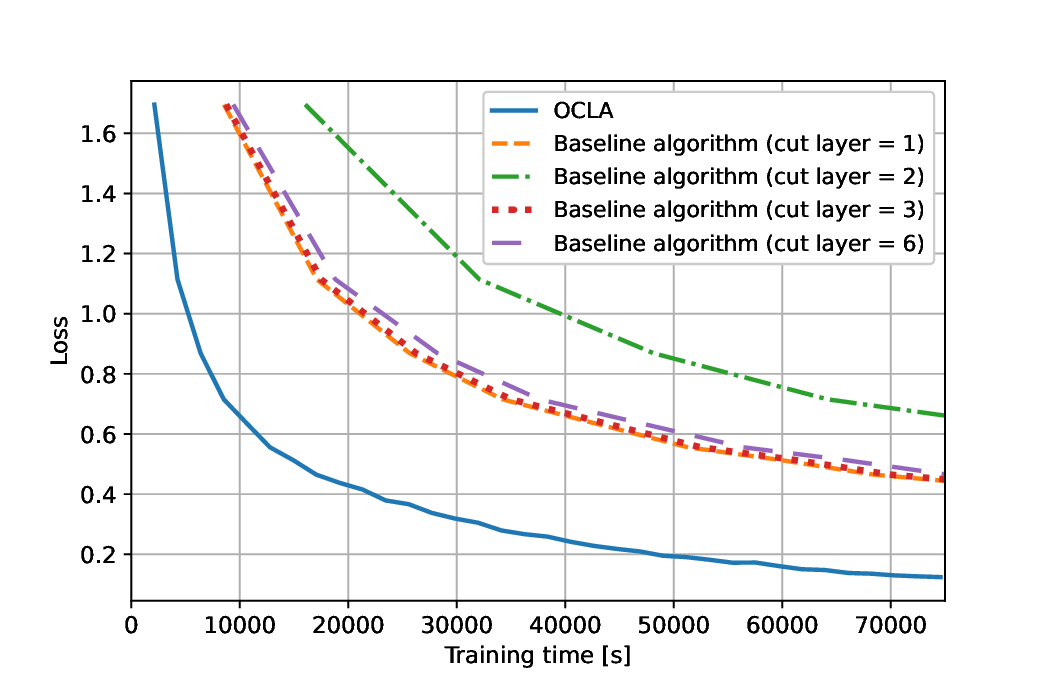} \vspace{-2mm}
\caption{Comparison between OCLA and the baseline algorithm selecting various cut layers in terms of the training loss vs. training time; $R_{cv}=0.5$ and $(1-\beta)_{cv}=0.5$} \vspace{-2mm}
\label{loss}
\end{figure}

The results illustrated in Fig. \ref{loss}, clearly show the importance of identifying a fitting cut layer in terms  of the optimization problem at hand. As demonstrated, OCLA converges much faster than the baseline algorithm. Specifically, the considerably different convergence rates when applying OCLA and the given benchmarks, are due to the different cut layer decisions that each of the algorithms establishes. As such, they result in dissimilar training delays, despite having the same hyperparameter configuration. It is evident that the proposed algorithm exhibits the highest convergence rate, since it consistently and successfully determines the optimal split layer. Conversely, Fig. \ref{loss} shows that each benchmark severely hinders the learning process by introducing high training delay caused by the frequently made suboptimal cut layer decisions.

\begin{figure}[t]
\vspace{-3mm}
\centering
\includegraphics[scale=0.49]{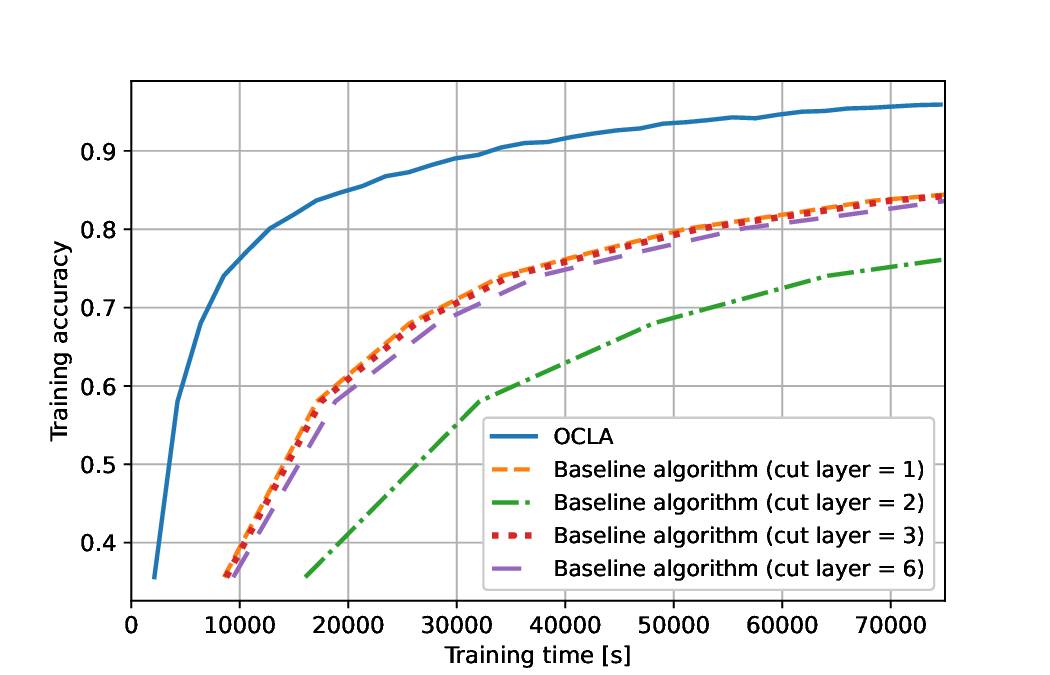} \vspace{-2mm}
\caption{Comparison between OCLA and the baseline algorithm selecting various cut layers in terms of training accuracy vs. training time; $R_{cv}=0.5$ and $(1-\beta)_{cv}=0.5$ \vspace{-1mm}}
\label{accuracy}
\end{figure}

In addition, the results depicted in Fig. \ref{accuracy} demonstrate that using OCLA also provides higher training accuracy and drastically accelerates the convergence of the model.

As indicated by these findings, merely configuring adequate hyperparameters does not lead to the desired model performance in practical scenarios that involve time-varying and resource-constrained systems. As such, optimal performance can only be achieved through considering the interrelation between the given optimization problem and the critical system properties, i.e. time-variability and resource limitations.

\section{Conclusion}

This study suggests that split learning-based control of EMG prosthetics is a viable and promising option. However, the cut layer decision in such resource-constrained and time-varying SL systems profoundly impacts many system-related aspects, including the model’s convergence rate. Hence, the paper develops an algorithm for identifying the position of the optimal cut layer with regard to convergence rate maximization, depending on the available system resources and the neural network architecture at hand. While the optimal cut layer drastically accelerates the learning, choosing a suboptimal one may lead to severe consequences regarding the convergence rate of the model. This conclusion implies that for ensuring the effectiveness of integrating SL in EMG-driven prosthetic control, the system resources and the deep learning model architecture must be considered.

\section*{Appendix A}

Within this appendix we clarify the reasoning behind the profile function pruning step.

The total communication overhead incorporates the transmission of activations/gradients and the parameters for weight synchronization. As such, when selecting layer $n$ as cut layer, the total communication overhead is given by $\frac{2D_k}{B_k}N_k(n)B_k+2N_c(n)$, where $\frac{D_k}{B_k}$ denotes the number of batches in the epoch, and $N_c(n)$ is the cumulative function of the parameters.

Considering that setting layer $n+1$ leads to higher computation delay, in order for it to remain in the pool of potential cut layers, layer $n+1$ needs to have lower transmission delay compared to layer $n$. Accordingly, it needs to have lower communication overhead as well. This leads to the conclusion that layer $n+1$ will be eliminated from the selection process if the following condition is satisfied: \vspace{-2mm}

\begin{small}
\begin{align}
\begin{array}{ll} \notag
\frac{2D_k}{B_k}N_k(n+1)B_k+2N_c(n+1) \geq 
\frac{2D_k}{B_k}N_k(n)B_k+2N_c(n) \vspace{+2mm}\\
N_k(n+1)D_k+N_c(n+1) \geq N_k(n)D_k+N_c(n),
\end{array}
\end{align}
\end{small}
which results in the concluding criterion for profile function pruning as specified in (\ref{comm}):

\begin{small}    
\begin{equation}
N_k(n+1)+\frac{N_c(n+1)-N_c(n)}{D_k} \geq N_k(n) \notag
\end{equation}
\end{small}

\section*{Appendix B}

This appendix presents the derivations of Lemma \ref{lemma1} and Lemma \ref{lemma2}.

\section*{Appendix B.1}
\section*{Proof of Lemma \ref{lemma1}}

The optimal cut layer $n$ must comply with condition (\ref{Tplus}). Following that, some elementary transformations are applied according to (\ref{totalDelay})-(\ref{tp}): \vspace{-2mm}

\begin{footnotesize}
\begin{align*}
&T(n+1) > T(n)\\
&\frac{2D_k}{B_k}[\tau_k(n+1)+t_0(n+1)+\tau_s(n+1)]+2t_p(n+1)\\
&>\frac{2D_k}{B_k}[\tau_k(n)+t_0(n)+\tau_s(n)]+2t_p(n)\\
&\frac{2D_k}{B_k}[\frac{L_k(n+1)B_k}{f_k}+\frac{N_k(n+1)B_k}{R}+\frac{L_s(n+1)B_k}{f_s}]+\frac{2Nc(n+1)}{R}\\
&>\frac{2D_k}{B_k}[\frac{L_k(n)B_k}{f_k}+\frac{N_k(n)B_k}{R}+\frac{L_s(n)B_k}{f_s}]+\frac{2N_c(n)}{R}\\
&D_k[\frac{L_k(n+1)}{f_k}+\frac{N_k(n+1)}{R}+\frac{L_s(n+1)}{f_s}]+\frac{N_c(n+1)}{R}\\
&>D_k[\frac{L_k(n)}{f_k}+\frac{N_k(n)}{R}+\frac{L_s(n)}{f_s}]+\frac{N_c(n)}{R}\\
&D_k[\frac{L_k(n+1)-L_k(n)}{f_k}+\frac{L_s(n+1)-L_s(n)}{f_s}]\\
&>D_k\frac{N_k(n)-N_k(n+1)}{R}-\frac{N_c(n+1)-N_c(n)}{R}\\
&D_k[\frac{L_k(n+1)-L_k(n)}{f_k}+\frac{L_{total}-L_k(n+1)-[L_{total}-L_k(n)]}{f_s}]\\
&>D_k\frac{N_k(n)-N_k(n+1)}{R}-\frac{N_c(n+1)-N_c(n)}{R}\\
&\frac{L_k(n+1)-L_k(n)}{f_k}-\frac{L_k(n+1)-L_k(n)}{af_k}\\
&>\frac{N_k(n)-N_k(n+1)}{R}-\frac{N_c(n+1)-N_c(n)}{D_kR} \\
&(a-1)\frac{L_k(n+1)-L_k(n)}{af_k}>\frac{N_k(n)-N_k(n+1)-\frac{N_c(n+1)-N_c(n)}{D_k}}{R}
\end{align*}
\end{footnotesize}
\vspace{-3mm}

In the previous steps, the substitution $a=\frac{f_s}{f_k}$ is used. By introducing another substitution regarding the computing speeds of the entities, $\beta=\frac{a-1}{a}$, the inequation undergoes further simplification: \vspace{-2mm}

\begin{footnotesize}
\begin{align}
&\frac{\beta[L_k(n+1)-L_k(n)]}{f_k}>\frac{N_k(n)-N_k(n+1)-\frac{N_c(n+1)-N_c(n)}{D_k}}{R}\\ \notag
&\frac{\beta R}{f_k}>\frac{N_k(n)-N_k(n+1)-\frac{N_c(n+1)-N_c(n)}{D_k}}{L_k(n+1)-L_k(n)}
\end{align}
\end{footnotesize}

The right side of the previous expression is defined by (\ref{gain_plus}), hence Lemma \ref{lemma1} is derived.

\section*{Appendix B.2}
\section*{Proof of Lemma \ref{lemma2}}

The optimal cut layer $n$ must comply with condition (\ref{Tminus}). Subsequently, similar elementary transformations are performed based on (\ref{totalDelay})-(\ref{tp}):

\balance
\begin{footnotesize}
\begin{align*}
&T(n-1) > T(n) \vspace{+1mm}\\
&\frac{2D_k}{B_k}[\tau_k(n-1)+t_0(n-1)+\tau_s(n-1)]+2t_p(n-1)\\
&>\frac{2D_k}{B_k}[\tau_k(n)+t_0(n)+\tau_s(n)]+2t_p(n)\\
&\frac{2D_k}{B_k}[\frac{L_k(n-1)B_k}{f_k}+\frac{N_k(n-1)B_k}{R}+\frac{L_s(n-1)B_k}{f_s}]+\frac{2N_c(n-1)}{R}\\
&>\frac{2D_k}{B_k}[\frac{L_k(n)B_k}{f_k}+\frac{N_k(n)B_k}{R}+\frac{L_s(n)B_k}{f_s}]+\frac{2N_c(n)}{R}\\
&D_k[\frac{L_k(n-1)}{f_k}+\frac{N_k(n-1)}{R}+\frac{L_s(n-1)}{f_s}]+\frac{N_c(n-1)}{R}\\
&>D_k[\frac{L_k(n)}{f_k}+\frac{N_k(n)}{R}+\frac{L_s(n)}{f_s}]+\frac{N_c(n)}{R}\\
&D_k\frac{N_k(n-1)-N_k(n)}{R}-\frac{N_c(n)-N_c(n-1)}{R}\\
&>D_k[\frac{L_k(n)-L_k(n-1)}{f_k}+\frac{L_s(n)-L_s(n-1)}{f_s}]\\
&D_k\frac{N_k(n-1)-N_k(n)}{R}-\frac{N_c(n)-N_c(n-1)}{R}\\
&>D_k[\frac{L_k(n)-L_k(n-1)}{f_k}+\frac{L_{total}-L_k(n)-[L_{total}-L_k(n-1)]}{f_s}]\\
&\frac{N_k(n-1)-N_k(n)}{R}-\frac{N_c(n)-N_c(n-1)}{D_kR}\\
&>\frac{L_k(n)-L_k(n-1)}{f_k}-\frac{L_k(n)-L_k(n-1)}{af_k}\\
&\frac{N_k(n-1)-N_k(n)-\frac{N_c(n)-N_c(n-1)}{D_k}}{R}>(a-1)\frac{L_k(n)-L_k(n-1)}{af_k}
\end{align*}
\end{footnotesize}

The derivation proceeds with substituting $\beta=\frac{a-1}{a}$, where $a=\frac{f_s}{f_k}$. The substitution results in the following simplified inequation: 

\begin{footnotesize}
\begin{align}
&\frac{N_k(n-1)-N_k(n)-\frac{N_c(n)-N_c(n-1)}{D_k}}{R}>\frac{\beta[L_k(n)-L_k(n-1)]}{f_k} \\ \notag
&\frac{N_k(n-1)-N_k(n)-\frac{N_c(n)-N_c(n-1)}{D_k}}{L_k(n)-L_k(n-1)}>\frac{\beta R}{f_k}
\end{align}
\end{footnotesize}

The left side of the previous expression is defined by (\ref{gain_minus}), thus the derivation of Lemma \ref{lemma2} is obtained.

\end{document}